%% file: main.tex
\documentclass[runningheads]{llncs}
\usepackage[T1]{fontenc}
\usepackage{placeins}
\usepackage{array}
\usepackage{multirow}
\usepackage{xcolor}%

\usepackage{graphicx} 
\usepackage{booktabs}
\usepackage[acronym]{glossaries} 
\glsdisablehyper
\loadglsentries{glossary.tex} 
\usepackage{graphicx,verbatim}
\begin{document}
%
\title{US-X Complete: A Multi-Modal Approach to Anatomical 3D Shape Recovery}
%

\author{Miruna-Alexandra Gafencu \inst{1,2,3} \and
Yordanka Velikova \inst{1,2} \and
Nassir Navab\inst{1} \and Mohammad Farid Azampour\inst{1,2}}

\authorrunning{M-A. Gafencu et al.}
%
\institute{Computer-Aided Medical Procedures (CAMP), Technical University of Munich, Munich, Germany \and Munich Center for Machine Learning (MCML), Germany
\and Konrad Zuse School of Excellence in Reliable AI (relAI), Germany}

\maketitle              
\begin{abstract}

Ultrasound offers a radiation-free, cost-effective solution for real-time visualization of spinal landmarks, paraspinal soft tissues and neurovascular structures, making it valuable for intraoperative guidance during spinal procedures. However, ultrasound suffers from inherent limitations in visualizing complete vertebral anatomy, in particular vertebral bodies, due to acoustic shadowing effects caused by bone. 
In this work, we present a novel multi-modal deep learning method for completing occluded anatomical structures in 3D ultrasound by leveraging complementary information from a single X-ray image. To enable training, we generate paired training data consisting of: (1) 2D lateral vertebral views that simulate X-ray scans, and (2) 3D partial vertebrae representations that mimic the limited visibility and occlusions encountered during ultrasound spine imaging. Our method integrates morphological information from both imaging modalities and demonstrates significant improvements in vertebral reconstruction (p < 0.001) compared to state of art in 3D ultrasound vertebral completion. We perform phantom studies as an initial step to future clinical translation, and achieve a more accurate, complete volumetric lumbar spine visualization overlayed on the ultrasound scan without the need for registration with preoperative modalities such as computed tomography.
This demonstrates that integrating a single X-ray projection mitigates ultrasound's key limitation while preserving its strengths as the primary imaging modality. Code and data can be found at https://github.com/miruna20/US-X-Complete


\end{abstract}
%
%
%


\section{Introduction}

Ultrasound imaging is increasingly gaining attention in spinal procedures, due to its ability to provide real-time, radiation-free visualization of paraspinal soft tissues and superficial bone landmarks. As an affordable and highly portable modality, ultrasound allows for bedside or operating-room use in diverse settings. In spinal injections such as epidural steroid injections, or lumbar facet joint injections the use of ultrasound increases the accuracy of needle placement and reduces the number of performed punctures at a lower radiation cost compared to fluoroscopy~\cite{rasoulian2015ultrasound,kalagara2021ultrasound}. In surgical procedures such as spinal tumor resections, ultrasound enables delineating the margins of tumors which minimizes the risk of residual tumor~\cite{zhou2011intraoperative}, while in spinal decompression procedures, ultrasound provides real time feedback of spinal cord and dural movements~\cite{kimura2012ultrasonographic}.
Overall, spine sonography not only provides real-time, radiation free guidance and assistance during these procedures, but also minimizes the risk for complications and surgical trauma. 

However, spinal ultrasound imaging has well known limitations and challenges. The bony structures of the spine severely attenuate ultrasound, causing acoustic shadows under the surface. These shadows occlude deeper structures such as the laminae, pedicles and the vertebral body~\cite{li2021image}. In practice, an acquisition only covers a few vertebral levels per sweep, giving a limited field of view relative to the full spine. These constraints mean that ultrasound images capture partial surface information and often only for the nearest surface. Moreover, image quality and interpretation is highly dependent on the skill of the operator. Inexperienced users may fail to obtain suitable scan planes, leading to inconsistent or incomplete bone visualization. Therefore, ultrasound alone does not readily offer an easily interpretable, volumetric view of spine anatomy which limits standalone ultrasound guidance~\cite{hu2025standalone}. 

To address the challenge of incomplete information, previous methods have explored registering intraoperative ultrasound images to preoperative CT scans. This enables clinicians to overlay accurately depicted CT-derived anatomy onto the live intraoperative ultrasound view, providing a more complete visualization of the spine. However, this process is technically challenging. 
Differences in patient positioning between the preoperative \gls{CT} and the intraoperative setting can lead to changes in spine curvature and, consequently, anatomical misalignments that rigid registration alone cannot resolve~\cite{azampour2024anatomy,nagpal2015multi,nagpal2014ct}. Moreover, patient movement during surgery typically requires the registration to be repeated. Finally, the approach depends on the availability of a recent CT scan, which may not always be feasible due to clinical, logistical or radiation-related constraints.    

As an alternative, Gafencu et al.~\cite{gafencu2024shape} proposed reconstructing the full spine anatomy directly and solely from intraoperative ultrasound. While this eliminates the need for preoperative imaging, it inherits the above-mentioned fundamental limitations of ultrasound, most notably the systematical occlusion of entire sub-structures of a vertebra such as the vertebral body. As a result, the inverse problem of reconstruction becomes highly under-constrained and ill-posed. Although this method produces anatomically plausible spine completions on both phantom and patient data, it struggles to estimate accurate vertebral body dimensions in real anatomical settings.

Given the limitations of both CT-based registration and ultrasound-only reconstruction, we propose a lightweight, intraoperative solution that combines ultrasound with a single lateral X-ray. This hybrid approach leverages the complementary strengths of the two modalities. A lateral radiograph offers a global, projection-based view of the full spinal column, capturing vertebral alignment and anchoring the overall geometry and scale. Ultrasound, on the other hand, provides localized, real-time information on bony landmarks and surrounding soft tissues. Acquiring a single X-ray is fast, introduces minimal radiation, and is already part of many spine procedures for level confirmation. By fusing these two data sources, we enable accurate, patient-specific 3D reconstruction of the spine without relying on preoperative imaging or complex intraoperative registration.

We formulate the recovery of the complete spine as a 3D shape completion problem. Point cloud-based shape completion has been widely studied in computer vision, particularly on synthetic object datasets such as ShapeNet~\cite{chang2015shapenet}. Early methods used encoder–decoder architectures such as PCN~\cite{yuan2018pcn} to reconstruct full shapes from partial inputs with extensions introducing hierarchical generation TopNet~\cite{tchapmi2019topnet} or parametric surface decoders~\cite{groueix2018papier}. Recent works explore advanced architectures such as transformed-based models like PoinTr~\cite{yu2021pointr} that capture long-range structural dependencies in the point set and methods such as  SnowflakeNet~\cite{xiang2021snowflakenet} that progressively refine details by fractally splitting points with skip-transformer blocks. Other methods, such as VRCNet~\cite{pan2021variational} use variational inference with relational reasoning to capture shape priors and local geometric structures more effectively.

Multi-modal, also called cross-modal shape completion extends this paradigm by integrating additional inputs to guide completion.  Typically, features from different modalities are fused during coarse shape prediction and subsequent refinement. For instance, ViPC~\cite{zhang2021view} leverages a single-view image for its global context to infer missing parts, while CSDN~\cite{zhu2023csdn} treats shape completion as a style transfer problem, using images to refine coarse predictions. Other methods incorporate attention mechanisms to fuse multi-view 2D and 3D information~\cite{Zhangetal2007}, or apply view-based reasoning for refinement. Beyond RGB guidance, additional modalities such as semantic segmentation~\cite{yang2021semantic}, temporal sequences~\cite{shi2022temporal}, textual descriptions~\cite{kasten2023point}, or geometric priors like symmetry~\cite{ren2022self} have been employed to further improve performance.
These methods often rely on aligned modalities and abundant training data with uniformly distributed occlusions. In contrast, medical imaging presents fundamentally different challenges: occlusions are structured and modality-specific, anatomical variability is high, and training data is limited.

In this work, we introduce the first multi-modal shape recovery framework designed specifically for spine imaging. Our approach integrates anatomical information from a 3D-tracked ultrasound(US) sweep and a single lateral X-ray by registering them into a shared 3D representation space to form a unified, multi-modal partial observation. We then reconstruct the full spine using a two-stage, coarse-to-fine probabilistic deep learning framework built upon VRCNet~\cite{pan2021variational} and enhance it with modality fusion modules. The final 3D reconstruction can be overlaid onto the ultrasound volume to support downstream navigation and interpretation. For training, we use simulated, modality-specific partial views. At inference time, our framework operates using only the 3D ultrasound scan and a single lateral X-ray, eliminating the need for CT/US registration. Our contributions are threefold:

(1) A novel multi-modal shape completion framework that combines X-ray and ultrasound for lumbar spine reconstruction, leveraging their complementary strengths and addressing the limitations of ultrasound-only approaches.\par
(2) A joint representation space that fuses anatomical features from both modalities, specifically designed for accurate shape reasoning.\par
(3) Comprehensive validation using synthetic data and physical phantom experiments conducted under operating room conditions. 

\section{Methodology}

\subsection{Data Generation}
\begin{figure*}[t]
    \centering
    \includegraphics[width=\textwidth, keepaspectratio]{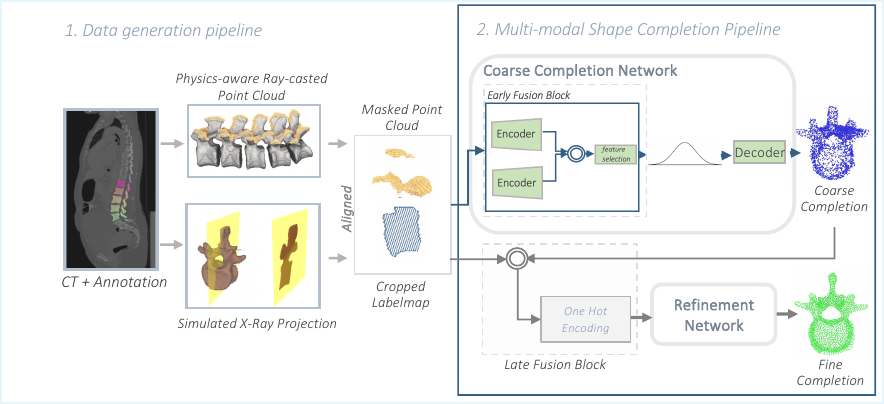}

    \caption{Overview of the proposed method: (a) Data Generation Pipeline: Synthetic training data is derived from annotated CT scans, simulating ultrasound-consistent partial vertebral observations to mimic acoustic shadowing. Simultaneously, 2D lateral X-ray projections of 3D vertebral segmentations are generated. These multi-modal observations are merged into an anatomically aligned 3D point cloud representation.(b) The aligned observations undergo a two-stage completion process. The coarse stage extracts global features from both modalities to generate a vertebral template, which is then refined with detailed features from ultrasound and X-ray data to produce the final complete shape.}
\label{fig:methodology}

\end{figure*}
Within an operating room setting for spine procedures, ultrasound scans enable real-time guidance of tools toward the target site, while X-ray imaging is typically used for final confirmation of their placement relative to the vertebral anatomy. To replicate this imaging scenario, we simulate X-ray and ultrasound views of the spine from the annotated CT scans of the VerSe2020~\cite{loffler2020vertebral} dataset. Displayed in Figure~\ref{fig:methodology}, our synthetic data generation accounts for the physical principles and acquisition constrains of each modality.
\subsubsection{Simulation of vertebral partial observation from ultrasound.}
Ultrasound imaging of the spine is fundamentally limited by acoustic shadowing. This effect renders large portions of the vertebrae invisible. As shown in the compounded 3D ultrasound scan displayed in Figure \ref{fig:phantom_validation}, typically only the vertebral arch is distinguishable. 

To simulate this partial visibility, we generate ultrasound-consistent point clouds from CT-derived spine meshes using a physics-aware ray-casting method~\cite{gafencu2024shape}. We simulate transverse ultrasound acquisitions by positioning a virtual camera above each spinous process and casting rays into the mesh. Only points with a surface normal forming an angle of less than 90° with the incoming ray are retained, mimicking the reflection behavior of ultrasound waves at oblique angles. This ensures that structures not favorably oriented toward the probe are appropriately occluded.

To further approximate ultrasound-specific artifacts such as off-plane scattering, we apply a set of lateral and anterior-posterior shifts to the mesh prior to ray-casting. By retaining only points that remain visible across these perturbed acquisitions, we simulate the effect of incoherent echo returns and spatial shadowing. The resulting surface point clouds exhibit varying degrees of occlusion and limited field of view, closely reflecting real ultrasound scans of the spine.

Finally, to generate inputs for vertebra-wise completion, we heuristically segment the spine into individual vertebral levels by applying fixed-size bounding box masks centered on each vertebral centroid. This results in noisy partial point clouds, which improve the completion network’s robustness to vertebral segmentation inaccuracies. This dataset has been used to train and evaluate the ultrasound-based shape completion method in ~\cite{gafencu2024shape}.

\subsubsection{Simulation of vertebral partial observation from X-ray.}
Lateral spinal X-ray scans, as shown in Figure~\ref{fig:phantom_validation}, capture vertebral bodies but lack depth information along the ray path. To simulate this projection, we start from the 3D vertebral segmentations in VerSe20 and project each vertebra's points onto the central transverse plane along the left–right axis, corresponding to the assumed X-ray beam direction in a lateral view. The resulting 2D projection and the projection plane (highlighted in yellow) are illustrated in Figure~\ref{fig:methodology}.
We then assign the third coordinate z of each point to the midpoint slice of the 3D segmented volume along the same axis, effectively placing the 2D projection into 3D space. This simulated 3D X-ray observation enables alignment with the corresponding 3D surface observations from ultrasound.

\subsubsection{Joint Representation Space.}\label{joint_rep_space}
Combining different imaging modalities in an anatomically-consistent manner remains a key challenge in medical imaging analysis~\cite{darzi2024review}. To address this, we construct a joint 3D point cloud representation that integrates anatomical information from both ultrasound and X-ray within a shared coordinate space.

In the simulated setting, the partial views are inherently registered, as both originate from the same CT scan. As a result, the simulated ultrasound surface (marked in orange) and the simulated X-ray projection expanded from 2D to 3D as previously explained (marked in blue) are geometrically and anatomically aligned, forming a coherent multi-modal observation. This observation is further used as input to the multi-modal shape completion pipeline.
 
In contrast, such inherent registration is not present in the OR. The registration strategy for the realistic OR scenario is described in Section~\ref{Subsection:phantom_validation}.

\subsection{Multi-modal Shape Completion Pipeline}\label{Subsection:multi-modal_network}
Our multi-modal shape completion pipeline, illustrated in Figure~\ref{fig:methodology} comprises two main stages: a coarse shape completion network that learns prior shape distribution of the vertebrae, and a refinement stage based on self-attention mechanisms that recovers detailed anatomical structures. Both stages are implemented as variational autoencoders and trained jointly using a combined loss function with two terms: \gls{KL} as a distribution divergence term and \gls{CD} term to supervise completion accuracy. To enable integration of complementary information from different imaging modalities, we introduce two novel components: an \emph{Early Fusion} block in the coarse completion stage and a \emph{Late Fusion} block in the refinement stage.

\subsubsection{Coarse Completion Stage} \par
\begin{figure*}[t]
    \centering
    \includegraphics[width=\textwidth, keepaspectratio]
{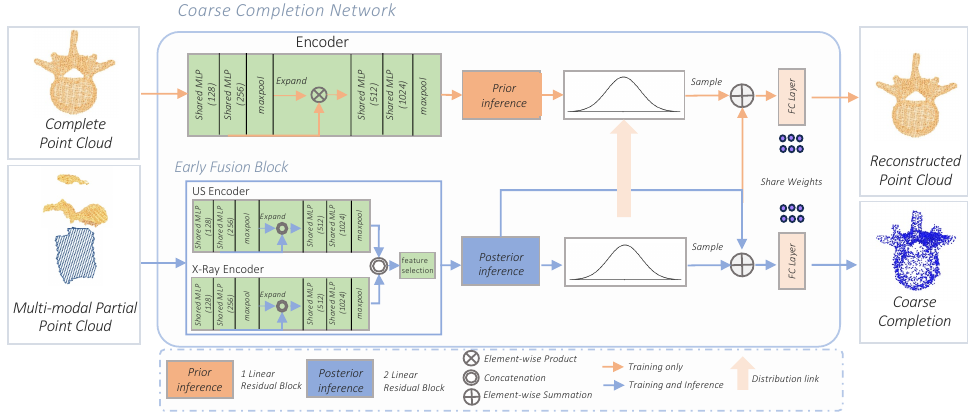}
    \caption{The coarse-stage network is trained to reconstruct the full vertebral shape from ground truth data, which implicitly teaches it a shape prior for lumbar vertebrae. Simultaneously, it learns to complete the shape conditioned on multi-modal partial observations from ultrasound and X-ray. As a result, the network outputs anatomically plausible coarse vertebral templates that integrate prior knowledge with image-based observations.  }
    \label{fig:coarsecompletionnetwork}
\end{figure*}

The coarse completion network depicted in Figure~\ref{fig:coarsecompletionnetwork} is trained to simultaneously reconstruct the full vertebral shape from complete input and to perform shape completion from partial observations. This dual objective enables the model to capture a prior distribution over complete vertebrae geometries while also learning to infer plausible completions from limited observations. 

To effectively integrate the complementary anatomical cues provided by ultrasound and X-ray segmentations, we introduce an Early Fusion module. Each modality is first processed by a dedicated MLP-based encoder that produces modality-specific latent features. These features are concatenated and passed through a feature selection block to project them into a unified 1024-dimensional latent space that serves as a shared representation across modalities. This representation is then passed to a posterior inference module, which estimates a Gaussian distribution over the latent space. During training, we enforce alignment between the posterior derived from partial inputs and the prior learned from complete shapes via a \gls{KL} divergence loss. Sampling from the prior enables the network to reconstruct full shapes, while sampling from the posterior enables coarse completion conditioned on partial observations. 

At inference time, only partial data is available. The model encodes the ultrasound and X-ray inputs, samples from the learned posterior, and generates a coarse completion that is then passed to the refinement stage. At this stage, the completed point cloud is predicted in the same coordinate space as the multi-modal partial observation and is therefore inherently co-registered to it. This property ensures that subsequent fusion and refinement operations can be performed without additional alignment steps. 

\subsubsection{Refinement Stage} 
To enhance the anatomical accuracy of the initial coarse prediction, we integrate a refinement stage using a point-based encoder–decoder architecture with integrated self-attention mechanisms as proposed by Pan et al.~\cite{pan2021variational}. This network is designed to aggregate point-level features across multiple spatial scales, which is particularly important for vertebral shape completion: vertebrae have complex morphological patterns with fine-grained details, such as processes, and curvature that are difficult to recover from global encodings and shape priors alone. 

To incorporate local geometric details from multi-modal partial observations into the refinement process, we introduce a late fusion strategy. In this setup, the coarse shape completion, the ultrasound  segmentation, and the X-ray segmentation are first concatenated into a single point cloud. Each point is then augmented at input stage with a one-hot encoding that defines its origin, allowing the network to distinguish between data sources.

This fusion strategy enables the network to treat each input differently. This distinction is critical, as the three sources differ fundamentally in nature: the coarse completion is a prior-based prediction, the X-ray point cloud is a projective 2.5D observation, and the ultrasound data, although noisy contains rich local morphological information. By explicitly encoding these differences we hypothesise that the network can effectively make use of the complementary strengths of each modality to improve refinement quality.

\subsection{Evaluation methodology}
We evaluate the impact of X-ray information as additional guidance to the ultrasound-based shape completion.  Furthermore, we investigate the role of integrating X-ray segmentation at different stages of our end-to-end shape completion pipeline within our ablation studies. This comprehensive analysis helps identify the optimal strategy for incorporating X-ray segmentation into the shape completion process. We additionally perform phantom studies and present both quantitative and qualitative results for a comprehensive understanding of our outcomes. 

Our evaluation combines established shape completion metrics such as \gls{CD}, \gls{EMD} and F1-score~\cite{gafencu2024shape} with anatomically-specific evaluation. To obtain additional insights, we perform a separate evaluation of the vertebral arch and body completion accuracy. Heuristically, we separate the structures based on the vertebra's centre of gravity \begin{math}cg \in R^3 \end{math}. Assuming anteroposterior axis alignment with the \begin{math} y \end{math} axis, we define \begin{math} Arch Points \in R^3, \end{math} \begin{math} ArchPoints=\{p| p_y > cg_y\} \end{math} for the vertebral arch and \begin{math} Body Points \in R^3\end{math}, \begin{math} BodyPoints=\{q| q_y <cg_y\}\end{math} for the vertebral body.

Statistical significance is assessed using the Wilcoxon signed-rank test, chosen for our paired, non-normally distributed results (confirmed by Shapiro-Wilk test).

\section{Experimental Setup}
\subsection{Image Acquisition Setup for Phantom Validation}\label{Subsection:phantom_validation}
\begin{figure*}[t]
    \centering
    \includegraphics[width=\textwidth, keepaspectratio]{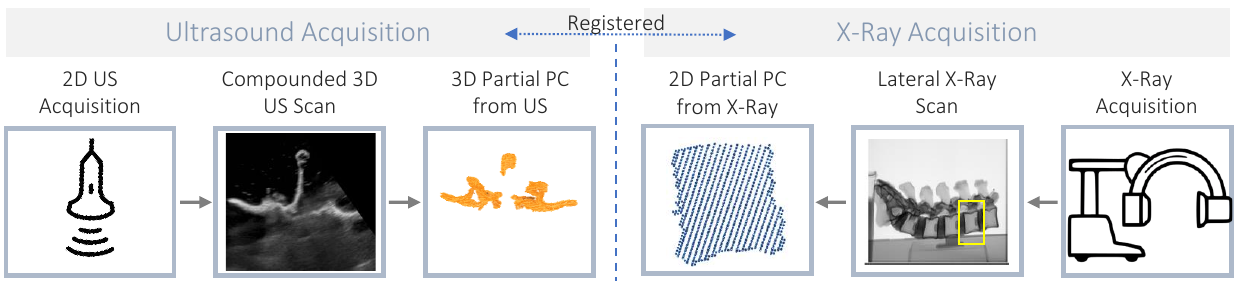}
    \caption{ Data acquisition and pre-processing for validating the proposed method using two spine phantoms. We acquire registered ultrasound and X-ray scans and integrate both modalities into our unified 3D point cloud representation before feeding it into the multi-model shape completion pipeline.}
    \label{fig:phantom_validation}
\end{figure*}

We conduct experiments on two spine phantoms (see appendix for details) containing lumbar vertebrae L1–L5, using a clinical-like setup to evaluate the feasibility of clinical translation of our proposed method (Figure~\ref{fig:phantom_validation}). Transverse ultrasound scans are acquired with a 2D curvilinear transducer (5C1) connected to an ACUSON Juniper ultrasound system\footnote{Siemens Healthineers, Germany}. The probe is mounted on the end-effector of a robotic manipulator\footnote{KUKA LBR iiwa 14 R820, KUKA Roboter GmbH, Augsburg, Germany} using a custom 3D-printed holder, and robotically tracked to enable 3D volume compounding. From the compounded volume, vertebrae are manually segmented and divided into individual levels.

We also acquire paired lateral X-ray and cone-beam \gls{CT} (CBCT) scans using a Loop-X system\footnote{medPhoton GmbH, Salzburg, Austria}, aligning the X-ray projection with the phantom’s left–right axis to capture lateral spine views. Since the system acquires both X-ray and CBCT images, they are inherently co-registered. We further register the CBCT volume to the ultrasound data using the registration approach described by Li et al.~\cite{li2025robotic}. We segment vertebral bodies semi-automatically from the X-ray scan and extract the complete ground truth spine shape from the CBCT, which we then divide into individual levels.

To place the X-ray scan within the 3D ultrasound volume, we first use the known registration chain between X-ray/CBCT and CBCT/ultrasound. This allows alignment along the craniocaudal and anteroposterior directions. To estimate the position of the X-ray scan along the lateral axis, we analyze the segmented spine in the ultrasound volume. Specifically, we compute the oriented bounding box of the segmented spine and determine its second principal axis, which typically corresponds to the left–right anatomical direction. Along this axis, we identify the two outermost points of the segmentation, representing the lateral extent of the visible spine, and calculate their midpoint. This midpoint serves as a reference for placing the X-ray scan in the left–right direction, ensuring it is centered relative to the spine and positioned in an anatomically meaningful way.

Finally, the registered ultrasound and X-ray segmentations are transformed into a multi-modal partial point cloud within the joint representation space described in Section~\ref{joint_rep_space}. This point cloud is subsequently passed to the shape completion network.

\subsection{Training Setup}
Throughout the experiments, a synthetic dataset derived from VerSe20 is used, split 60-20-20 into training, validation, and testing sets, totaling 149 lumbar spine samples, each containing 5 vertebrae.
Training is conducted over 100 epochs using the Adam optimizer (learning rate = 0.0001), with batch sizes of 4 and 2 for training and testing, respectively.
All experiments are performed on an NVIDIA GeForce RTX 4080 GPU. At inference time, completing one vertebra, on average, takes 0.31 seconds.

\section{Results}
\begin{table}[]
\centering
\caption{Quantitative performance comparison (in terms of \gls{CD}, \gls{EMD} multiplied by $10^{4}$ and F1-Score) of our proposed multi-modal method compared to the ultrasound-based baseline~\cite{gafencu2024shape}  on synthetic and phantom data, as well as results of the ablation studies of Early Fusion(EF) and Late Fusion(LF).}
    \begin{tabular}{llcccccc} \toprule
    \multirow{2}{*}{\textbf{Method}} & \multirow{2}{*}{ } 
    & \multicolumn{3}{c}{\textbf{Synthetic}} 
    & \multicolumn{3}{c}{\textbf{Phantom}} \\ 
    \cmidrule(lr){3-5} \cmidrule(lr){6-8}
    & & {\textbf{CD$\downarrow$}} & {\textbf{EMD$\downarrow$}} & {\textbf{F1-Score$\uparrow$}} 
    & {\textbf{CD$\downarrow$}} & {\textbf{EMD$\downarrow$}} & {\textbf{F1-Score$\uparrow$}} \\ 
    \midrule
    \multirow{2}{*}{Baseline~\cite{gafencu2024shape}} & Arch & 5.3$\pm$1.8 & 329.5$\pm$65.2 & 0.34$\pm$0.06 & 11.4$\pm$2.4 & 647.4$\pm$85.2 & 0.24$\pm$0.03 \\  
                               & Body & 7.7$\pm$4.8 & 349.2$\pm$90.3 & 0.30$\pm$0.10 & 20.7$\pm$5.1 & 530.4$\pm$46.6 & 0.14$\pm$0.03 \\  
    \midrule
    \multirow{2}{*}{\textbf{Ours}} & Arch & \textbf{4.6$\pm$1.4} & \textbf{279.4$\pm$48.1} & \textbf{0.41$\pm$0.07} & \textbf{7.8$\pm$2.5} & \textbf{422.6$\pm$59.1} & \textbf{0.35$\pm$0.05} \\  
                                    & Body & \textbf{4.0$\pm$1.6} & \textbf{252.8$\pm$74.8} & \textbf{0.48$\pm$0.10} & \textbf{7.1$\pm$2.0} & \textbf{359.3$\pm$79.5} & \textbf{0.32$\pm$0.05} \\  
    \midrule
    \multirow{2}{*}{EF} & Arch & 5.3$\pm$1.8 & 317.4$\pm$57.1 & 0.32$\pm$0.07 & 10.9$\pm$2.9 & 491.7$\pm$54.9 & 0.19$\pm$0.05 \\  
                                   & Body & 8.0$\pm$4.7 & 360.4$\pm$103.9 & 0.27$\pm$0.09 & 22.7$\pm$6.8 & 569.8$\pm$83.7 & 0.14$\pm$0.03 \\  
    \midrule
    \multirow{2}{*}{LF} & Arch & 5.6$\pm$2.4 & 303.7$\pm$56.3 & 0.36$\pm$0.07 & 10.8$\pm$5.0 & 444.8$\pm$74.0 & 0.31$\pm$0.06 \\  
                                  & Body & 4.2$\pm$2.3 & 263.8$\pm$64.3 & 0.46$\pm$0.10 & 13.7$\pm$9.5 & 386.1$\pm$81.4 & 0.29$\pm$0.05 \\  
    \bottomrule
    \end{tabular}
\label{table:results}
\end{table}

\begin{figure*}[t]
    \centering
    \includegraphics[width=\textwidth, keepaspectratio]
{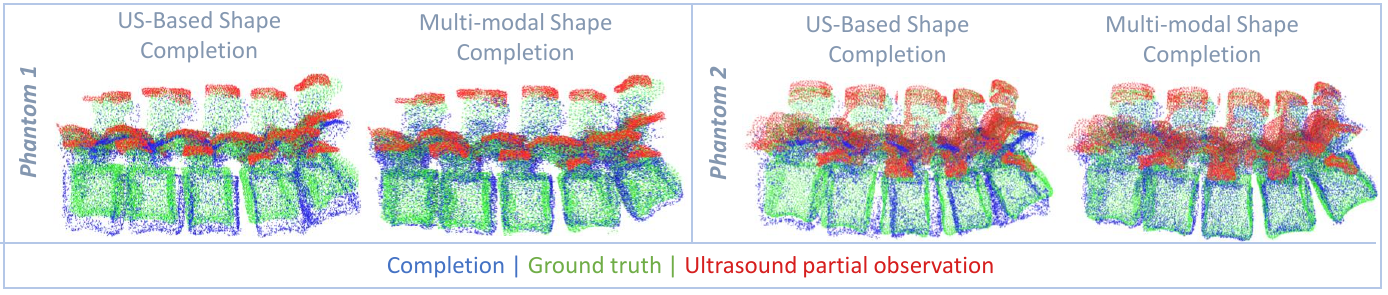}
    \caption{Qualitative comparison of previously introduced ultrasound-based vertebral shape completion method~\cite{gafencu2024shape} versus our proposed multi-modal approach on two spine phantoms.}
    \label{fig:qualitative_results_phantom_exp1}
\end{figure*}

\subsection{Evaluation of the proposed method against baseline}
We evaluate our multi-modal shape completion approach against previously introduced baseline that uses only partial observations from ultrasound~\cite{gafencu2024shape}. Table~\ref{table:results} displays results on synthetic and phantom dataset. 
From the synthetic dataset evaluation we demonstrate that our multi-modal approach significantly improves vertebral shape completion across all metrics with p-values smaller than 1.00e-6. 

The phantom evaluation further validates these findings, with substantial improvements in vertebral body accuracy compared to the baseline with a mean \gls{CD} difference of 13.6. Figure~\ref{fig:qualitative_results_phantom_exp1} provides visual evidence of these improvements through a lateral view of the completed spine phantoms, specifically showing how a single X-ray scan helps correct both vertebral body shape and size. This shows that our network, trained on synthetic data, successfully generalizes to the phantom-based clinical-like scenarios and effectively integrates real X-ray data.

Importantly, our network is trained exclusively on synthetic data but generalizes effectively to phantom datasets, demonstrating strong robustness to domain shifts and validating its potential for clinical transfer. Furthermore, we observe consistent improvements not only in the vertebral body but also in the vertebral arch. This suggests that the availability of global context from the X-ray data may provide indirect constraints that help refine the geometry of adjacent structures.

\subsection{Ablation studies}\label{Subsection:experiments_fusiontype}
In this experiment, we conduct ablation studies to evaluate the impact of different architectural design choices for multi-modal shape completion, as introduced in Section~\ref{Subsection:multi-modal_network}. Table~\ref{table:results} summarizes the results, with rows labeled EF (Early Fusion) and LF (Late Fusion) and average scores over all 140 completed vertebrae from the synthetic dataset and a total of 10 vertebrae from the two lumbar phantoms.  

Integrating X-ray segmentation through late fusion results in a significant improvement in shape completion accuracy across all metrics (p-values < 1.00e-6). This suggests that incorporating X-ray morphological information during the refinement stage effectively captures the geometric details specific to the projective modality.

In contrast, stand-alone early fusion has a more limited effect. While the encoded global representation of the X-ray segmentation slightly improves the initial vertebral template in the coarse stage, it does not substantially enhance the final output. Combining early and late fusion achieves the highest overall accuracy, demonstrating that first providing a multi-modal global representation, followed by refinement with detailed geometric information, yields the most effective multi-modal shape completion.

\section{Discussion and Conclusion}
Our experimental results highlight the capability of our method to generalize effectively from synthetic datasets to phantom-based acquisitions. This successful generalization demonstrates that our synthetic data pipeline realistically captures the characteristics of both ultrasound and X-ray imaging modalities, suggesting strong potential for translation to clinical patient data. Further improvements could be performed by additionally accounting for diverse patient positioning and therefore various spine curvatures to improve generalizibility across different setups. 

To fully benefit from this capability in clinical settings, our approach requires precise alignment between ultrasound and X-ray modalities. In our current setup, the placement of the X-ray within the ultrasound volume is heuristically determined based on the spatial extent of the ultrasound spine segmentation. This heuristic approach can be affected by anatomical variability, limited ultrasound field of view, and segmentation noise, potentially introducing alignment errors. Although ultrasound/X-ray registration introduces its own set of challenges, intraoperatively acquired X-ray images naturally reflect the patient’s real-time posture, reducing misalignment compared to preoperative modalities such as CT. Nevertheless, even small registration inaccuracies can lead to erroneous shape estimations, representing a current limitation of our method. To address this, future work should focus on improving the robustness of the shape completion pipeline against such misalignments or exploring registration-free alternatives, while still leveraging global geometric cues to guide the reconstruction process effectively.

Looking forward, increasing shape completion accuracy could be achieved by incorporating anatomical constraints across the entire spinal structure. Currently, our reconstruction operates vertebra-by-vertebra. However, treating the spine as an interconnected structure with inherent spatial constraints between adjacent vertebrae could lead to more accurate and anatomically consistent results.

In conclusion, this study represents the first exploration of multi-modal medical imaging for shape completion in spinal procedures. By integrating critical vertebral information from ultrasound and X-ray imaging, our approach overcomes the inherent limitations of standalone ultrasound visualization. This initial proof-of-concept integration within the clinical workflow lays the foundation for automated ultrasound-based navigation and guidance and represents a further step towards intelligent, anatomy-aware surgical guidance systems.
\FloatBarrier

\appendix
\section{Appendix}
\subsection{Phantom Description}
\begin{figure*}
    \centering
    \includegraphics[width=\textwidth, keepaspectratio]{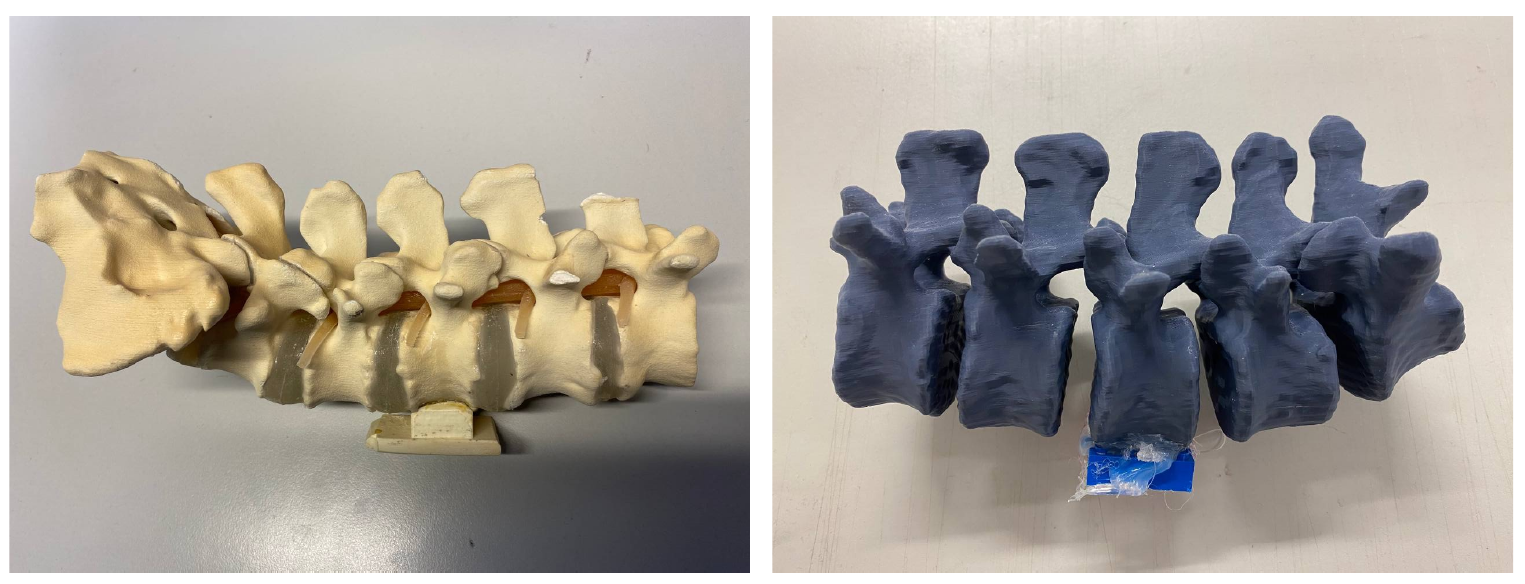}
    \caption{Two lumbar spine phantoms utilized to conduct validation of our method within a clinical-like setup.}
\label{fig:phantoms}
\end{figure*}

We have conducted experiments using two lumbar spine phantoms displayed in Figure~\ref{fig:phantoms}. The first phantom, displayed on the left side, contains the five lumbar vertebrae L1-L5 and the sacrum as well as intervertebral disks from a different, softer material. The second phantom, displayed on the right side, is a 3D printed version of the volumetric annotations of lumbar vertebrae from subject 818 in the VerSe 2020 dataset. 
\FloatBarrier




%
%
%
\bibliographystyle{splncs04}
\bibliography{refs}

\end{document}

%% file: glossary.tex

\newacronym[shortplural={D$_{\text{dye}}$}, longplural={donor dye, ex. Alexa 488}]{ddye}{D$_{\text{dye}}$}{donor dye, ex. Alexa 488}

\newacronym{US}{ultrasound}{}
\newacronym{DL}{DL}{deep learning}
\newacronym{CT}{CT}{computer tomography}
\newacronym{MRI}{MRI}{magnetic resonance imaging}

\newacronym{STOA}{STOA}{state-of-the-art}
\newacronym{TRE}{TRE}{target registration error}
\newacronym{OR}{OR}{operating room}
\newacronym{FL}{FL}{x-ray fluoroscopy}
\newacronym{CAD}{CAD}{computer-aided design}
\newacronym{AR}{AR}{augmented reality}
\newacronym{VR}{VR}{virtual reality}
\newacronym{MLP}{MLP}{multi-layer perceptron}
\newacronym{SDF}{SDF}{signed distance function}
\newacronym{PCN}{PCN}{Point Completion Network}
\newacronym{MSN}{MSN}{Morphing and Sampling Network}
\newacronym{FinerPCN}{FinerPCN}{Finer Point Completion Network}
\newacronym{ViPC}{ViPC}{View-Guided Point Cloud Completion}
\newacronym{PointCNN}{PointCNN}{Point Convolutional Neural Network}
\newacronym[plural=CNNs,firstplural=convolutional neural networks (CNNs)]{CNN}{CNN}{convolutional neural network}
\newacronym[plural=IFNs,firstplural=Implicit Feature Networks]{IFN}{IFN}{Implicit Feature Network}
\newacronym{PPE}{PPE}{Point Positional Encoding}
\newacronym{GRNet}{GRNet}{Gridding Residual Network }
\newacronym{VE-PCN}{VE-PCN}{Voxel Edge Point Completion Network}
\newacronym{MRAC-Net}{MRAC-Net}{Multi-resolution Anisotropic Convolutional Network}
\newacronym[plural=VAEs, firstplural= Variational Autoencoders (VAEs)]{VAE}{VAE}{Variational Autoencoder}
\newacronym[plural=GANs, firstplural= Generative Adversarial Networks (GANs)]{GAN}{GAN}{generative adversarial network}
\newacronym{AE}{AE}{auto encoder}
\newacronym{PCT}{PCT}{Point Completion Transformer}
\newacronym{SA-Net}{SA-Net}{Shuffle Attention Network}
\newacronym{DSC}{DSC}{dice score}
\newacronym{HD}{HD}{Hausdorf Distance}
\newacronym{VerSe20}{VerSe20}{Large Scale Vertebrae Segmentation Challenge Dataset 2020}
\newacronym{PMNet}{PMNet}{Probabilistic Modeling Network}
\newacronym{RENet}{RENet}{Relational Enhacement Network }
\newacronym{CD}{CD}{Chamfer Distance}
\newacronym{EMD}{EMD}{Earth Mover's Distance}
\newacronym{PC}{PC}{point cloud}

\newacronym{F1}{F1}{F1-score}
\newacronym{VRCNet}{VRCNet}{Variational Relational Completion Network}
\newacronym{GT}{GT}{ground truth}
\newacronym{SOTA}{SOTA}{state-of-the-art}
\newacronym{KL}{KL}{Kullback-Leibler}
\newacronym{CL}{CL}{Chamfer Loss}
\newacronym{CBCT}{CBCT}{cone beam CT}
\newacronym{AP}{AP}{anterior-posterior}

\newacronym{R-PSK}{R-PSK}{Residual Point Selective Kernel Module}
\newacronym{FC}{FC}{Fully Connected}